\documentclass[conference]{IEEEtran}
\IEEEoverridecommandlockouts
\usepackage{amsmath,amssymb,amsfonts}
\usepackage{algorithmic}
\usepackage{graphicx}
\usepackage{textcomp}
\usepackage{csquotes}
\usepackage{xcolor,soul}
\usepackage{adjustbox}
\usepackage{hyperref}
\usepackage{multirow}
\usepackage{titlesec}
\usepackage{subcaption}
\usepackage{makecell}
\usepackage{orcidlink}

\usepackage[numbers,sort&compress]{natbib}

\usepackage{etoolbox}

\definecolor{columbiablue}{rgb}{0.61, 0.87, 1.0}
\titlespacing{\subsubsection}{0pt}{0pt}{1pt}

\usepackage{amsmath}

\DeclareMathOperator*{\argmin}{arg\,min}
\usepackage{mathtools}
\usepackage[colorinlistoftodos,prependcaption,textsize=tiny]{todonotes}
\usepackage{gensymb}
\usepackage{multirow}

\usepackage{parskip}

\begin{document}

\title{A Data-Driven Method for INS/DVL Alignment}
\author{\IEEEauthorblockN{Guy Damari\orcidlink{ 
0009-0001-6394-6026} and Itzik Klein\orcidlink{0000-0001-7846-0654}}
\IEEEauthorblockA{{The Hatter Department of Marine Technologies} \\
{Charney School of Marine Sciences, University of Haifa}\\
Haifa, Israel}
}

\maketitle
\begin{abstract}
Autonomous underwater vehicles (AUVs) are sophisticated robotic platforms crucial for a wide range of applications. The accuracy of AUV navigation systems is critical to their success. Inertial sensors and Doppler velocity logs (DVL) fusion is a promising solution for long-range underwater navigation. However, the effectiveness of this fusion depends heavily on an accurate alignment between the inertial sensors and the DVL. While current alignment methods show promise, there remains significant room for improvement in terms of accuracy, convergence time, and alignment trajectory efficiency. In this research we propose an end-to-end deep learning framework for the alignment process. By leveraging deep-learning capabilities, such as noise reduction and capture of nonlinearities in the data, we show using simulative data, that our proposed approach enhances both alignment accuracy and reduces convergence time beyond current model-based methods.
\end{abstract}

\section{Introduction}\label{intro_sec}
Autonomous underwater vehicles (AUVs) have revolutionized our understanding of the world's oceans, playing a vital role in marine research, industry, and exploration. A critical factor underpinning the success of AUV missions is the accuracy and reliability of their navigation systems \cite{kinsey2006survey}. Unlike aerial or terrestrial vehicles, AUVs must navigate in an environment where global navigation satellite system (GNSS) technology is unavailable \cite{liu2018innovative}. This absence of global positioning reference presents a fundamental challenge for underwater navigation, necessitating alternative methods for accurate position determination. \\
\noindent 
To address this challenge, the integration of inertial navigation systems (INS) and Doppler velocity logs (DVL) has emerged as one of the most promising methods for precise long-range navigation of AUVs. An INS typically comprises a three-axis accelerometer and a three-axis gyroscope. The accelerometer measures the AUV's specific force vector, while the gyroscope measure the angular velocity vector \cite{groves2015principles}. Together, these sensors enable the INS algorithm to calculate the vehicle's position, velocity, and orientation in three-dimensional space without external references. This system provides highly accurate short-term position data but suffers from cumulative errors over time due to inherent sensor biases and noise \cite{titterton2004strapdown}. Consequently, relying solely on INS for underwater navigation is impractical \cite{akeila2013reducing}. \\
\noindent
DVL, on the other hand, employs the Doppler effect to measure the vehicle's velocity relative to the seafloor. The DVL employs a configuration of four acoustic transducers, each emitting a distinct beam toward the seafloor. By analyzing the reflected acoustic signals from the seafloor, the DVL provides precise measurements of the vehicle's velocity \cite{brokloff1994matrix,farrell2008aided}. The synergy between INS and DVL creates a robust navigation solution \cite{wang2019novel}. The INS provides high-frequency updates and handles rapid maneuvers well, while the DVL offers drift-free velocity measurements. This combination allows for more accurate dead reckoning, enabling AUVs to navigate with greater precision over extended periods.\\
\noindent
However, the effectiveness of this integration heavily depends on the accurate alignment between the inertial measurement unit (IMU), the core of the INS, and the DVL \cite{jalving2004dvl, troni2010new, whitcomb1999advances}. The alignment problem essentially involves estimating the transformation matrix between two coordinate systems: the IMU frame and the DVL frame. This challenge is closely related to Wahba's problem \cite{wahba1965least} in attitude determination, which seeks to find the optimal rotation matrix between two coordinate systems. Precise alignment is crucial because even small misalignments can lead to significant navigation errors over time, potentially jeopardizing mission objectives or the safety of the AUV \cite{jalving2004dvl}. \\
\noindent
Standard approaches to solving this alignment problem have shown adequate performance under certain conditions. These methods often involve specific alignment routines performed before deployment, such as prescribed motion patterns or the use of external reference systems. In \cite{kinsey2007situ}, the authors proposed a method utilizing long base-line (LBL) acoustic positioning information combined with least-square techniques to estimate the alignment for three degrees of freedom. Although the results were favorable, this approach, like other absolute acoustic navigation methods, relies on external beacons and sensors, which introduces both operational complexity and logistical challenges to AUV deployments.\\
\noindent
Troni et al. \cite{troni2012field} developed two novel alignment approaches: a velocity-based and an acceleration-based method, both relying exclusively on the vehicle's onboard INS and DVL sensors. Their experimental results demonstrated robust performance of these methods under actual deep-sea conditions, despite the presence of real sensor noise. Yet, these approaches have limitations: the achievable accuracy could be improved, and the methods demand both specific vehicle trajectories and extended convergence times. \\
\noindent
Advancements in machine learning, particularly in deep learning, offer new possibilities for addressing complex sensor fusion and alignment problems. Deep learning models have demonstrated remarkable capabilities in extracting meaningful patterns from high-dimensional data and learning complex non-linear relationships.
Recently, deep learning approaches were successfully applied to AUV navigation \cite{cohen2024inertial,he2023deep} including topics of DVL accuracy enhancement \cite{cohen2022beamsnet} missing beam construction \cite{yona2024missbeamnet} and uncertainty estimation in INS/DVL fusion \cite{cohen2025adaptive,levy2025adaptive}. \\
\noindent
Motivated by the achievements made in deep-learning underwater navigation, in this research we propose AligNet, a novel data-driven approach to IMU-DVL alignment using deep learning techniques. By leveraging large datasets of sensor measurements collected under various operational conditions, deep learning approaches have the potential to learn robust alignment models that can adapt to different scenarios and
potentially perform online realignment during missions. Our initial goal is focused on overcoming the limitations of current methods by: (1) Improving alignment accuracy across various operational conditions; (2) Reducing the time required for convergence to optimal alignment parameters, enabling more efficient pre-mission preparations; and (3) Minimizing the dependence on specific alignment trajectory patterns, thus increasing flexibility in real-world deployments. \\
\noindent
To evaluate our proposed approach, we developed a comprehensive simulation framework that generates realistic AUV trajectories with varying sensor error characteristics, including different levels of IMU and DVL noise and error terms. Using this framework, we generated a dataset of over 10,000 lawn mower trajectories to train and validate our deep learning model. We show that AlignNet achieves superior performance compared to the standard velocity-based method, reducing the required convergence time by 75\% while maintaining comparable accuracy. Furthermore, our approach demonstrates robust performance across different sensor error configurations, maintaining consistent alignment accuracy even when tested with IMU error characteristics significantly different from the training conditions. \\
\noindent
The rest of this paper is organized as follows: Section \ref{prob_form_sec} gives the mathematical formulation of the alignment problem. Section \ref{prop_approach} describes our proposed data-driven alignment method, AlignNet. Section \ref{res_sec} presents the results and Section \ref{conc_sec} concludes this paper.
\section{Problem Formulation}\label{prob_form_sec}
\subsection{DVL Velocity Estimation}\label{dvl_basic_eq}
The DVL transmits and receives acoustic beams to and from the ocean floor. By utilizing the Doppler shift effect, the DVL is able to estimate the AUV velocity in the DVL frame, denoted as $\boldsymbol{v}_{AUV}^{d}$. The beam arrays are generally mounted using the $"+"$ or $"\times"$ configurations \cite{liu2018innovative}. In the $"\times"$ configuration, also known as the "Janus" configuration, the beams are horizontally orthogonal with direction vectors defined by  \cite{cohen2022beamsnet, braginsky2020correction}:
\begin{equation}\label{bi_in_h}
    \centering
        \boldsymbol{b}_{\dot{\imath}}=
        \begin{bmatrix} 
        \cos{\psi_{\dot{\imath}}}\sin{\alpha}\quad
        \sin{\psi_{\dot{\imath}}}\sin{\alpha}\quad
        \cos{\alpha}
    \end{bmatrix}_{1\times3}
\end{equation}
where, $\psi_{\dot{\imath}}$ and $\alpha$ are the yaw and pitch angles of beam $i = 1,2,3,4$, respectively. For all beams, the pitch angle remains constant, whereas yaw angle differ for each beam as \cite{yona2021compensating}:
\begin{equation}\label{yaw_of_beams}
    \centering
        \psi_{\dot{\imath}}=(\dot{\imath}-1)\cdot\frac{\pi}{2}+\frac{\pi}{4}\;[rad]\;,\; \dot{\imath}=1,2,3,4
\end{equation}
Stacking all beam directions (1) gives:
\begin{equation}\label{H_mat_defenition}
    \centering
        \mathbf{H}=
        \begin{bmatrix} \boldsymbol{b}_{1}\\\boldsymbol{b}_{2}\\\boldsymbol{b}_{3}\\\boldsymbol{b}_{4}\\
    \end{bmatrix}_{4\times3}
\end{equation}

Using $\mathbf{H}$, the velocity vector of the AUV in the DVL frame is connected to the measured velocity vector of the beams \cite{cohen2022beamsnet}:
\begin{equation}\label{simple_H_to_v_dvl}
    \centering
    \boldsymbol{v}_{beams} = \mathbf{H}\boldsymbol{v}^{d}_{AUV}
\end{equation}
where $\boldsymbol{v}_{beams} \in \mathbb{R}^4$ is the beams velocity vector and $\boldsymbol{v}^{d}_{AUV} \in \mathbb{R}^3$ is the AUV velocity vector in the DVL frame. To emulate real world conditions, an error model is applied to each beam measurement \cite{tal2017inertial}:
\begin{equation}\label{y_as_func_of_H_and_v_error_model}
    \centering
        \boldsymbol{\tilde{y}} = [\mathbf{H} \boldsymbol{v}_{auv}^{d}(1+\boldsymbol{s}_{DVL})]+\boldsymbol{b}_{DVL}+\boldsymbol{\sigma}_{DVL}
\end{equation}
where $\boldsymbol{s}_{DVL}$ is the DVL's scale factor, $\boldsymbol{b}_{DVL}$ is the DVL's bias  and $\boldsymbol{\sigma}_{DVL}$ is the DVL's Gaussian distributed zero mean white noise.

The AUV velocity vector is obtained by minimizing the following cost:
\begin{equation}\label{ls_form_eq}
    \centering
    \tilde{\boldsymbol{v}}_{AUV}^{d} = \underset{\boldsymbol{{v}}_{AUV}^{d}}{\argmin}{||\boldsymbol{y}-\mathbf{H}\boldsymbol{{v}}_{AUV}^{d} ||}^{2}
\end{equation}
yielding a least squares solution:
\begin{equation}\label{psudo_inverse}
    \centering
    \tilde{\boldsymbol{v}}_{auv}^{d} = (\mathbf{H}^{T}\mathbf{H})^{-1}\mathbf{H}^{T}\boldsymbol{y}
\end{equation}

\subsection{INS Equations of Motion}\label{ins_equations}
The INS process the accelerometer and gyroscope measurements to estimate the velocity vector required for the INS/DVL alignment process.
Working with low-cost inertial sensors, certain simplifications can be made to the navigation equations of motion. Specifically, the Earth's rotation rate and transport rate effects can be neglected. Under these assumptions, the fundamental INS equations of motion are given by \cite{titterton2004strapdown}:
\begin{equation}
    \centering
    \dot{\boldsymbol{p}}^{n} = \boldsymbol{v}^{n}
\end{equation}
\begin{equation}
    \centering
    \dot{\boldsymbol{v}}^{n} = \mathbf{R}^{n}_{b}\boldsymbol{f}^{n}_{ib} + \boldsymbol{g}^{n}
\end{equation}
\begin{equation}
    \centering
    \dot{\mathbf{R}}^{n}_{b} = \mathbf{R}^{n}_{b}\boldsymbol{\Omega}^{b}_{ib}
\end{equation}
where ${\boldsymbol{f}}^{n}_{ib}$ is the specific force vector expressed in the navigation frame, $\boldsymbol{p}^{n}$ is the position vector expressed in the navigation frame, $\boldsymbol{v}^{n}$ is the velocity vector expressed in the navigation frame, $\mathbf{R}^{n}_{b}$ is the transformation matrix from body to navigation frame, $\boldsymbol{g}^{n}$ is the gravity vector expressed in the navigation frame, and $\boldsymbol{\Omega}^{b}_{ib}$ is the skew-symmetric matrix formed from the gyroscope measurements, defined as:
\begin{equation}
    \centering
    \boldsymbol{\Omega}^{b}_{ib} = 
    \begin{bmatrix}
    0 & -\omega_z & \omega_y \\
    \omega_z & 0 & -\omega_x \\
    -\omega_y & \omega_x & 0
    \end{bmatrix}
\end{equation}

To emulate real-world conditions we add commonly used error terms and noises to the inertial readings \cite{groves2015principles}.
The measured specific force vector is:
\begin{equation}
    \centering
    \tilde{\boldsymbol{f}}^{b}_{ib} = \boldsymbol{S}_a\boldsymbol{f}^{b}_{ib} + \boldsymbol{b}_a + \boldsymbol{\sigma}_a
\end{equation}
where $\tilde{\boldsymbol{f}}^{b}_{ib}$ is the measured specific force vector, $\boldsymbol{S}_a$ is a diagonal scale factor matrix, $\boldsymbol{b}_a$ is the bias vector, and $\boldsymbol{\sigma}_a$ is zero-mean Gaussian white noise. \\
\noindent
Similarly, for the angular rate measurements:
\begin{equation}
    \centering
    \tilde{\boldsymbol{\omega}}^{b}_{ib} = \boldsymbol{S}_g\boldsymbol{\omega}^{b}_{ib} + \boldsymbol{b}_g + \boldsymbol{\sigma}_g
\end{equation}
where $\tilde{\boldsymbol{\omega}}^{b}_{ib}$ is the measured angular velocity, $\boldsymbol{S}_g$ is the scale factor matrix, $\boldsymbol{b}_g$ is the bias vector, and $\boldsymbol{\sigma}_g$ is zero-mean Gaussian white noise.

\subsection{INS/DVL Alignment Problem}\label{ins_dvl_alignment}
The INS/DVL alignment problem involves determining the transformation matrix $\mathbf{R}_{b}^{d}$ between the body frame (b) and the DVL frame (d), assuming the inertial senors sensitive axes aligns with the body frame. This transformation is crucial for accurate fusion of INS and DVL measurements. The relationship between the velocities in these frames can be expressed as:
\begin{equation}\label{velocity_transformation}
    \centering
    \boldsymbol{\tilde{v}}^{d} = \mathbf{R}_{b}^{d} \boldsymbol{\tilde{v}}^{b}
\end{equation}
where $\boldsymbol{v}^{d}$ is the measured velocity vector expressed in the DVL frame and $\boldsymbol{v}^{b}$ is the inertial calculated velocity vector expressed in the the body frame.


\subsection{Model-based Alignment Method}\label{curr_alignment_methods}
The standard approach to INS/DVL alignment relies on velocity-based methods that integrate acceleration data from the IMU and compare it with velocity measurements from the DVL. This relationship can be expressed as \cite{troni2012field}:
\begin{equation}\label{velocity_based_eq}
    \centering
    \mathbf{R}_{i}^{b}(t)^{T}\int_{t_0}^{t}\mathbf{R}_{i}^{b}(\tau)\boldsymbol{a}^{b}(\tau)d\tau + \boldsymbol{v}^{b}(t_0) = \mathbf{R}_{b}^{d}\boldsymbol{v}^{d}(t)
\end{equation}
where $\mathbf{R}_{i}^{b}$ is the rotation matrix from the inertial frame to the body frame, $\boldsymbol{a}^{b}$ is the acceleration in the body frame, and $\boldsymbol{v}^{b}(t_0)$ is the initial velocity in the body frame.
The unknown rotation matrix $\mathbf{R}_{b}^{d}$ is typically estimated using the singular value decomposition (SVD) approach.



\section{Proposed Approach}\label{prop_approach}
While the SVD method has proven effective for certain applications, it
typically requires specific trajectory patterns and extended convergence
times, which can be operationally limiting for AUV deployments. These limitations motivate the development of our approach, AlignNet. It is a deep learning framework designed to estimate the alignment parameters between the INS and DVL reference frames using only the synchronized velocity measurements from both sensors. That is, we keep the same input for the alignment and replace the SVD approach with our AlingNet. \\
We next describe our network architecture and training process. Our approach offers advantages over standard methods, including faster convergence time, improved accuracy, and reduced dependence on specific trajectory patterns.




\subsection{Network Architecture}\label{network_architecture}
\begin{figure*}[!h]
    \centering
    \includegraphics[width=1\linewidth]{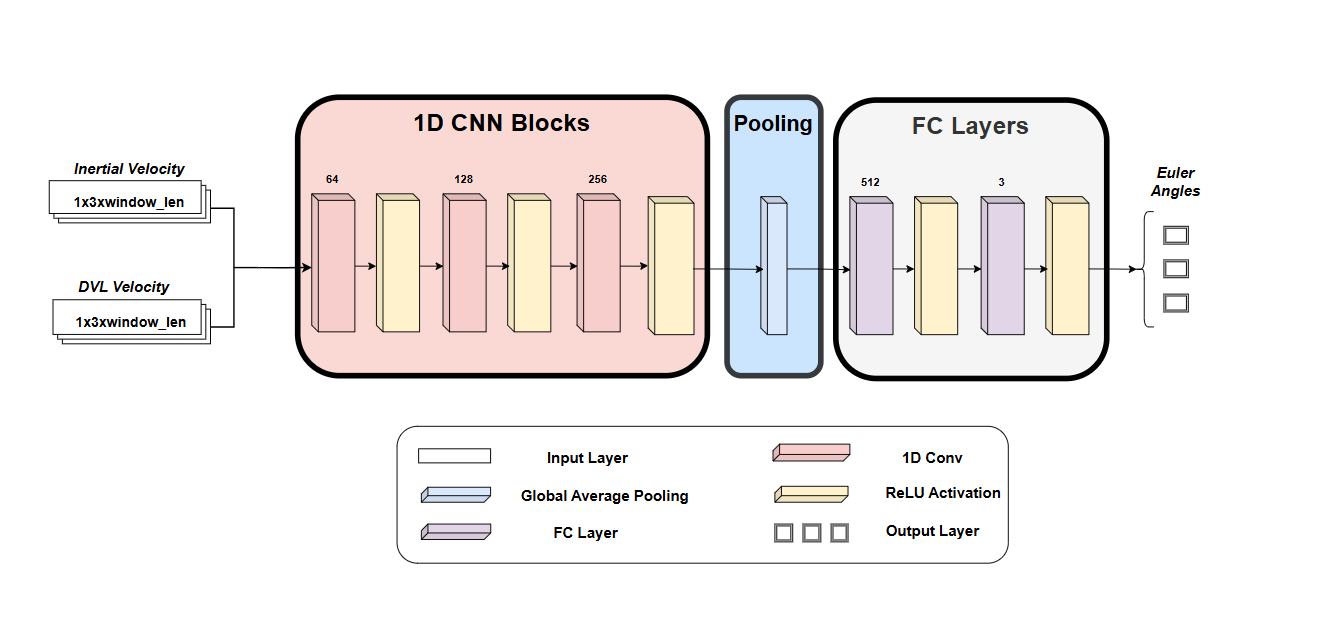}
    \caption{AlignNet architecture showing the processing pipeline from input velocity measurements to estimated Euler angles. The network consists of three main components: input preprocessing, 1D CNN blocks with increasing feature dimensions (64, 128, 256), a global average pooling layer, and fully connected layers that output the final alignment angles.}
    \label{fig:alignnet_architecture}
\end{figure*}

AlignNet employs a one-dimensional convolutional neural network (1D-CNN) architecture specifically designed for processing temporal sensor data from IMU and DVL measurements. The network's design enables it to automatically extract meaningful features from time series data while maintaining computational efficiency. Figure \ref{fig:alignnet_architecture} illustrates the complete architecture of AlignNet. \\
\noindent
The input layer accepts synchronized measurements from both IMU and DVL sensors, combining them into a six-dimensional tensor that captures the complete motion characteristics of the vehicle. The input tensor has dimensions $[batch\_size, window\_size, 6]$, where $window\_size$ represents the temporal dimension of the measurement window. \\
\noindent
The convolutional backbone consists of multiple 1D convolutional blocks with increasing feature dimensions (64, 128, and 256 channels) as shown in Figure \ref{fig:alignnet_architecture}. Each block contains convolutional layers with ReLU activation functions, implementing the core feature extraction process. These layers progressively learn hierarchical representations from the temporal sensor data, automatically identifying patterns relevant for alignment estimation. \\
\noindent
Following the convolutional layers, a global average pooling layer condenses the learned features across the temporal dimension. The pooled features are then processed through fully connected layers (with 512 neurons in the intermediate layer) that capture complex relationships in the data. The final output layer produces three values corresponding to the Euler angles (roll, pitch, and yaw) that define the rotation between the IMU and DVL reference frames.

\subsection{Training Process}\label{loss_function}
\begin{figure}[!h]
    \centering
    \includegraphics[width=1\linewidth]{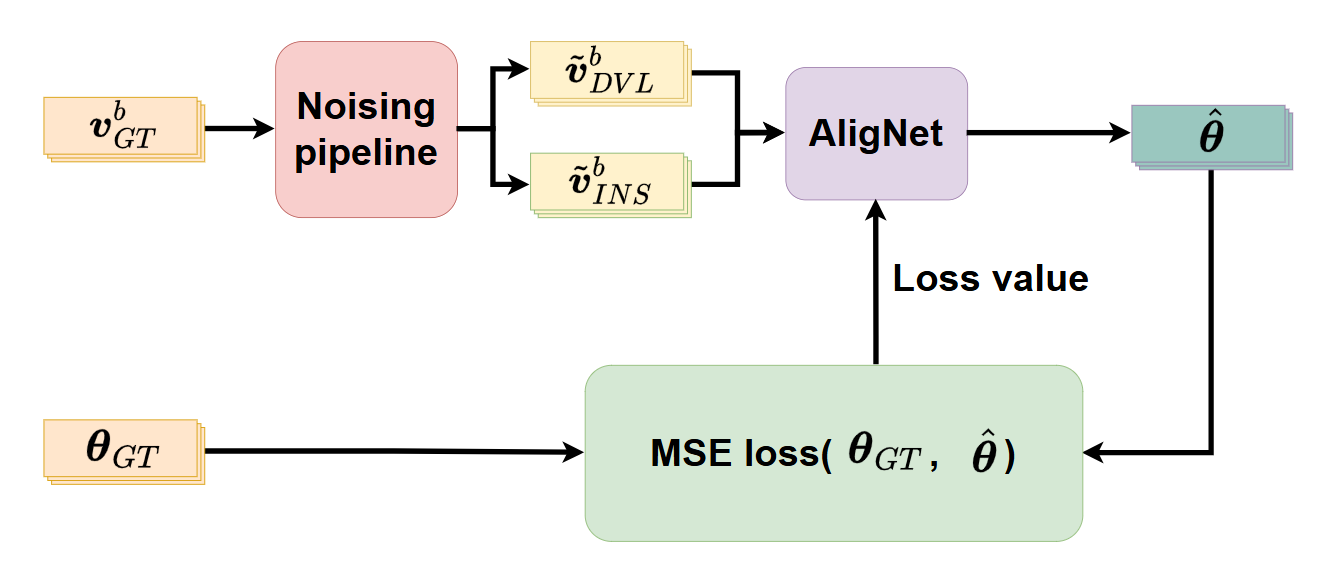}
    \caption{Overview of the AlignNet training pipeline. The ground truth velocity ($v^b_{GT}$) is processed through a noising pipeline to generate simulated DVL and INS velocities, which serve as input to AlignNet. The network estimates the alignment angles ($\hat{\theta}$), which are compared against the ground truth alignment angles ($\theta_{GT}$) using MSE loss to train the network.}
    \label{fig:alignet_flow}
\end{figure}

Figure \ref{fig:alignet_flow} presents the overall pipeline of our proposed approach.
The pipeline begins with ground truth velocity data that is processed through a noising pipeline to simulate realistic sensor measurements. These noisy measurements from both DVL and INS are then fed into AlignNet, which estimates the alignment angles between the sensor frames. The network is trained by comparing these estimated angles against ground truth alignment angles using mean squared error (MSE) loss, with the loss value feeding back to optimize the network parameters. \\
\noindent
The MSE is calculated across the three Euler angles that define the rotation between IMU and DVL frames:
\begin{equation}\label{rmse_eq}
    \centering
    \text{MSE}(\boldsymbol{\alpha}_{i} , \hat{\boldsymbol{\alpha}_{i}}) = \frac{1} {N}\sum_{i=1}^{N} \sum_{j = x,y,z}(\boldsymbol{\alpha}_{i,j} - \hat{\boldsymbol{\alpha}}_{i,j})^{2}
\end{equation}
\noindent
where $\alpha$ represents the ground truth Euler angles, $\hat{\alpha}$ represents the estimated Euler angles from AlignNet, N is the number of test samples, and j indexes the three rotational components (roll $\phi$, pitch $\theta$, and yaw $\psi$). \\
\noindent
The network is trained using the Adam optimizer with an initial learning rate of $10^{-7}$ and a batch size of 32. To prevent overfitting, we implement early stopping with a patience of 15 epochs and learning rate decay, reducing the rate by a factor of 0.5 when validation loss plateaus.


\section{Simulative Results}\label{res_sec}
\subsection{Simulation Setup}\label{simulation_setup}
\begin{figure}[!h]
    \centering
    \includegraphics[width=1\linewidth]{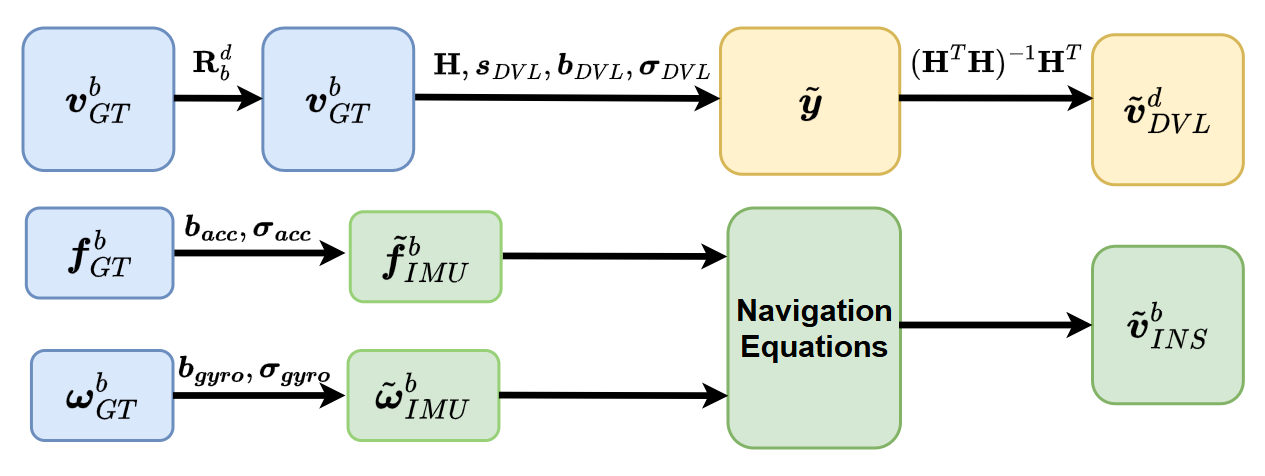}
    \caption{Simulation pipeline for generating DVL and INS velocity measurements. The upper path shows the DVL velocity generation process, where ground truth velocity is transformed and corrupted with noise and bias. The lower path shows the INS velocity computation through integration of noisy IMU measurements.}
    \label{fig:fig_sim_pipeline}
\end{figure}

We developed a comprehensive MATLAB-Simulink simulation framework to evaluate our proposed approach against the model-based baseline. The simulation incorporates a six-degree-of-freedom (6-DOF) AUV dynamics model with realistic hydrodynamic forces, detailed sensor error models, and various trajectory patterns. Figure \ref{fig:fig_sim_pipeline} illustrates the complete simulation pipeline for generating both DVL and INS velocity measurements.\\
\noindent
The simulation framework begins with the generation of the GT trajectory data, including position, velocity, and orientation parameters over time. These GT values serve as the baseline for evaluating alignment performance.
For sensor error modeling, we implement comprehensive error characteristics for both DVL and INS measurements. The DVL error model incorporates scale factors ($s_{DVL}$), biases ($b_{DVL}$), and Gaussian distributed measurement noise ($\sigma_{DVL}$). Similarly, the INS error model includes biases, and noise for both accelerometer and gyroscope measurements.
As shown in Figure \ref{fig:fig_sim_pipeline}, the simulation pipeline generates two parallel streams of velocity measurements. The upper path represents the DVL measurement process, where GT velocity undergoes coordinate transformation before being corrupted by the modeled errors. The lower path represents the INS velocity computation, where GT specific force and angular velocity measurements are first corrupted by sensor errors and then processed through the navigation equations to produce velocity estimates. \\
\noindent
The simulation parameters were configured to match the
physical characteristics of the "Snapir" AUV operated by the University of Haifa, including its mass,
inertia tensor, and hydrodynamic coefficients.
We generated synthetic datasets for training and testing with the following characteristics:
\begin{itemize}
    \item \textbf{Dataset Generation:} 10648 lawn mower trajectories were simulated, with each trajectory spanning 230 seconds. The IMU was sampled at 100 Hz while the DVL was sampled at 5 Hz, reflecting realistic sensor sampling rates in AUV systems.
    
    \item \textbf{Dataset Division:} The complete dataset was divided into training (60\%), validation (20\%), and testing (20\%) sets to ensure proper model evaluation and to prevent overfitting.
\end{itemize}
The IMU-DVL alignment configurations were systematically varied during dataset generation, with Euler angles covering a range of 0-45 degrees for each axis relative to the baseline configuration. This wide range of alignment variations ensures that the model learns to handle diverse misalignment scenarios that might be encountered in real-world AUV deployments.
\subsection{Performance Evaluation}\label{performance_evaluation}
\begin{figure}[!h]
    \centering
    \includegraphics[width=1\linewidth]{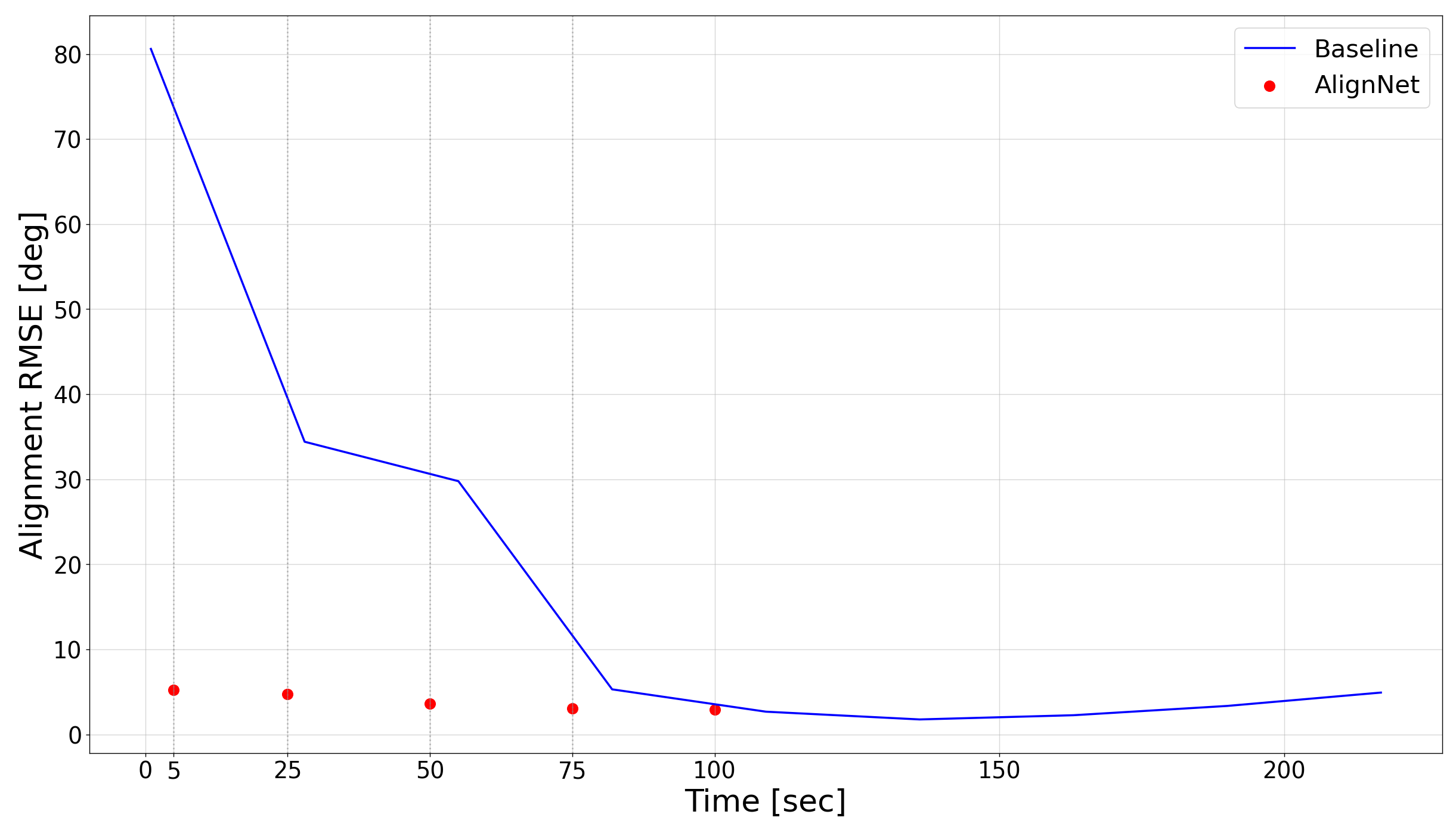}
    \caption{Comparison of alignment RMSE performance between AlignNet and the baseline velocity-based SVD alignment method over time on simulated lawn mower trajectory (230 seconds). AlignNet reaches comparable accuracy within 25 seconds that the baseline method requires 100 seconds to achieve.}
    \label{fig:rmse_comparison}
\end{figure}

We evaluated the performance of AlignNet against the baseline velocity-based SVD approach \cite{troni2012field} using a simulated lawn mower trajectory pattern, each with a duration of 230 seconds. The primary metrics for evaluation were alignment accuracy is root mean square error (RMSE) of Euler angles and the convergence time of the approach.\\
\noindent
Figure \ref{fig:rmse_comparison} shows the comparison of alignment RMSE performance between AlignNet and the baseline velocity-based SVD alignment method over time on this simulated lawn mower pattern dataset. 
The results show that AlignNet consistently improves upon the baseline method in terms of both accuracy and convergence time when evaluated on the lawn mower pattern. With just 5 seconds of data, AlignNet achieves an RMSE of 5.24°, compared to the baseline method's substantially higher error of 74.31°. AlignNet requires only 25 seconds to reach an RMSE of 4.72°, whereas the baseline method needs approximately 100 seconds to reach comparable accuracy (3.54°), representing a 75\% reduction in convergence time. At 100 seconds, AlignNet achieves an RMSE of 2.93°, demonstrating consistent improvement with additional data collection.
\subsection{Sensitivity to Sensor Noise}\label{noise_sensitivity}
To evaluate the robustness of AlignNet to different levels of inertial sensor noise and biases, we conducted experiments with varying error characteristics for the IMU sensors. The network was initially trained on simulated lawn mower trajectories with an IMU error model consisting of accelerometer bias of 100 {\micro}g and gyroscope bias of 1 deg/hour, with sensor noise levels set two orders of magnitude below these bias values. We then tested the network on four different IMU error configurations to assess how changes in sensor characteristics affect alignment performance. Table \ref{tab:noise_sensitivity} summarizes the alignment RMSE results for both AlignNet and the baseline velocity-based SVD method across different sensor error configurations and window sizes.\\
\noindent
Several key observations can be drawn from these results. First, AlignNet demonstrates remarkable consistency across different error configurations, especially at early time windows. For example, with just 25 seconds of data, AlignNet's RMSE remains between 2.72° and 4.72° across all configurations, while the baseline method's performance varies dramatically from 12.29° to 39.63°.
Second, while the baseline method eventually achieves superior accuracy with high-quality sensors (particularly with lower gyroscope bias of 0.1 °/h) and longer window sizes, it requires significantly more time to converge. This highlights that traditional methods can excel when provided with high-quality sensors and sufficient time, but AlignNet offers much more consistent performance across varying sensor qualities and time constraints.
Third, AlignNet shows better generalization to sensor characteristics different from its training data. Even when tested on IMU error models with twice the accelerometer bias (200 {\micro}g) or one-tenth the accelerometer bias (10 {\micro}g) compared to its training configuration, the network maintains stable performance.
\begin{table}[!h]
\caption{Alignment RMSE (degrees) for different accelerometer and gyroscope biases.}
\label{tab:noise_sensitivity}
\centering
\footnotesize
\setlength{\tabcolsep}{3pt}
\begin{tabular}{|c|c|cccc|}
\hline
Method & Time & \multicolumn{4}{c|}{IMU Error Config.} \\ 
\cline{3-6} 
 & & 100 {\micro}g & 200 {\micro}g & 10 {\micro}g & 10 {\micro}g \\
 & (s) & 1 °/h & 1 °/h & 1 °/h & 0.1 °/h \\ 
\hline
SVD & 5 & 74.31 & 72.36 & 74.62 & 73.71 \\ 
(baseline) & 25 & 39.63 & 37.20 & 34.65 & 12.29 \\ 
 & 50 & 30.61 & 29.53 & 27.35 & 4.04 \\ 
 & 75 & 11.62 & 13.17 & 8.90 & 1.49 \\ 
 & 100 & 3.54 & 5.55 & 2.04 & 0.41 \\ 
\hline
AlignNet & 5 & 5.24 & 3.58 & 3.78 & 3.48 \\ 
(ours) & 25 & 4.72 & 2.81 & 2.85 & 2.72 \\ 
 & 50 & 3.64 & 4.11 & 3.54 & 2.17 \\ 
 & 75 & 3.05 & 2.84 & 3.15 & 2.53 \\ 
 & 100 & 2.93 & 2.35 & 1.93 & 1.84 \\ 
\hline
\end{tabular}
\end{table}
%
\section{Conclusions}\label{conc_sec}
This paper presented AlignNet, a deep learning framework for INS/DVL alignment in autonomous underwater vehicles. Our approach leverages convolutional neural networks to automatically learn the optimal alignment between inertial sensors and DVL reference frames from synchronized velocity measurements.\\
\noindent
Simulation results demonstrate that AlignNet improves upon standard model-based alignment methods in terms of both accuracy and convergence time. AlignNet consistently achieves lower alignment errors with shorter data collection periods, while the baseline velocity-based SVD method requires substantially more time to reach comparable accuracy levels, representing a significant reduction in convergence time.
Furthermore, AlignNet shows consistent performance across different IMU error characteristics, maintaining stable alignment accuracy even when tested with sensor error models different from its training configuration. This robustness to sensor quality variations suggests potential applications in scenarios with cost or operational constraints.\\
\noindent
The primary advantages of AlignNet include faster convergence, as the data-driven approach determines alignment parameters with significantly fewer measurements than analytical methods; consistency across various sensor quality levels, offering stable performance where standard methods may struggle; and generalization ability to adapt to sensor characteristics different from its training conditions.\\
\noindent
Future work will focus on validating AlignNet using sea recorded data from AUV deployments in real-world conditions, testing the model's robustness against environmental factors and dynamic motion patterns. 
\section*{Acknowledgments}
G. D. is grateful for the support of the Maurice Hatter foundation and University of Haifa Data Science Research Center.



\bibliographystyle{IEEEtran}
\bibliography{References}

\end{document}